\pdfoutput=1

\documentclass[11pt]{article}

\usepackage[table]{xcolor}
\usepackage[final]{acl}
\usepackage{booktabs}

\usepackage{times}
\usepackage{latexsym}
\usepackage{amsmath}
\usepackage{graphicx}
\usepackage{amsmath}
\usepackage{amsfonts}
\usepackage{dsfont}
\usepackage{bbm}
\usepackage{xspace}
\usepackage{multirow}
\usepackage{csquotes}
\usepackage{refcount}
\usepackage{hyperref}

\usepackage[T1]{fontenc}

\usepackage[utf8]{inputenc}

\usepackage{microtype}
\usepackage{makecell}

\RequirePackage[normalem]{ulem} 
\RequirePackage{color}\definecolor{RED}{rgb}{1,0,0}\definecolor{BLUE}{rgb}{0,0,1} 
\providecommand{\DIFadd}[1]{{\protect\color{blue}\uwave{#1}}} 
\providecommand{\DIFdel}[1]{{\protect\color{red}\sout{#1}}}                      
\providecommand{\DIFaddbegin}{} 
\providecommand{\DIFaddend}{} 
\providecommand{\DIFdelbegin}{} 
\providecommand{\DIFdelend}{} 

\newcommand{\datawiki}{W\textsc{iki}D\textsc{omains}\xspace}
\newcommand{\scidef}{S\textsc{ci}D\textsc{ef}\xspace}
\newcommand{\difficultconcept}{{\it difficult concept}\xspace}
\newcommand{\domain}{{domain}\xspace}
\newcommand{\definition}{{\it definition}\xspace}
\newcommand{\concept}{{\it term}\xspace}

\newcommand{\taskname}{targeted concept simplification\xspace}

\newcommand{\tasknameemph}{{\it targeted} concept simplification\xspace}

\newcommand{\simplify}{$simplify$\xspace}
\newcommand{\explain}{$explain$\xspace}

\newcommand{\hmp}{$\mathcal{H}_{\textsc{mp}}$\xspace}
\newcommand{\hru}{$\mathcal{H}_{\textsc{ru}}$\xspace}
\newcommand{\hre}{$\mathcal{H}_{\textsc{re}}$\xspace}
\newcommand{\bleu}{$\textsc{Bleu}$\xspace}
\newcommand{\bleufour}{$\textsc{Bleu-4}$\xspace}
\newcommand{\bertscore}{$\textsc{BERTScore}$\xspace}

\definecolor{darkgreen}{HTML}{069631}

\title{Evaluating LLMs for Targeted Concept Simplification for Domain-Specific Texts}

\author{
  Sumit Asthana$^{\dagger}$\Thanks{Work done as student researcher at Google DeepMind.} \hspace{0.3cm} Hannah Rashkin$^\ddagger$\hspace{0.3cm} Elizabeth Clark$^\ddagger$ \\
  \textbf{Fantine Huot}$^\ddagger$\hspace{0.3cm} \textbf{Mirella Lapata}$^\ddagger$ \\
  $^\dagger$University of Michigan, Ann Arbor \quad $^\ddagger$Google DeepMind \\
  \texttt{asumit@umich.edu} \quad \texttt{\{hrashkin, eaclark, fantinehuot, lapata\}@google.com}
}

\begin{document}
\maketitle
\begin{abstract}
One useful application of NLP models is to support people in reading complex text from unfamiliar \domain{s} (e.g., scientific articles). Simplifying the entire text makes it understandable but sometimes removes important details. On the contrary, helping adult readers understand difficult concepts in context can enhance their vocabulary and knowledge. In a preliminary human study, we first identify that lack of context and unfamiliarity with difficult concepts is a major reason for adult readers' difficulty with domain-specific text. We then introduce \tasknameemph, a simplification task for rewriting text to help readers comprehend text containing unfamiliar concepts.  We also introduce \datawiki\footnote{\href{https://github.com/google-deepmind/wikidomains}{https://github.com/google-deepmind/wikidomains}}, a new dataset of 22k definitions from 13 academic \domain{s} paired with a \difficultconcept within each definition. We benchmark the performance of  open-source and commercial LLMs, and a simple dictionary baseline on this task across human judgments of ease of understanding and meaning preservation. Interestingly, our human judges preferred explanations about the \difficultconcept more than simplification of the concept phrase. Further, no single model achieved superior performance across all quality dimensions, and automated metrics also show low correlations with human evaluations of concept simplification ($\sim0.2$), opening up rich avenues for research on personalized human reading comprehension support.
\end{abstract}

\section{Introduction}
Text simplification helps lay audiences understand challenging text by simplifying difficult terms, syntax, or discourse~\cite{zhang-lapata-2017-sentence,agrawal-carpuat-2023-controlling} or by adding content to elaborate on the text~\cite{srikanth-li-2021-elaborative}.
With advances in neural models, especially LLMs, sentence simplification has made considerable progress towards generating text at different reading grade levels~\cite{kew-etal-2023-bless}. However, skilled adult readers face more challenges with lack of subject-matter knowledge~\cite{Guo2023PersonalizedJI}. Supporting readers in understanding concepts they find personally difficult within a larger body of text not only expands their vocabulary, but also helps them  develop a broader understanding of the topic~\cite{kintsch1991role,van2010using}.

\begin{figure}
    \centering
    \includegraphics[width=\linewidth, trim=12 10 380 0, clip]{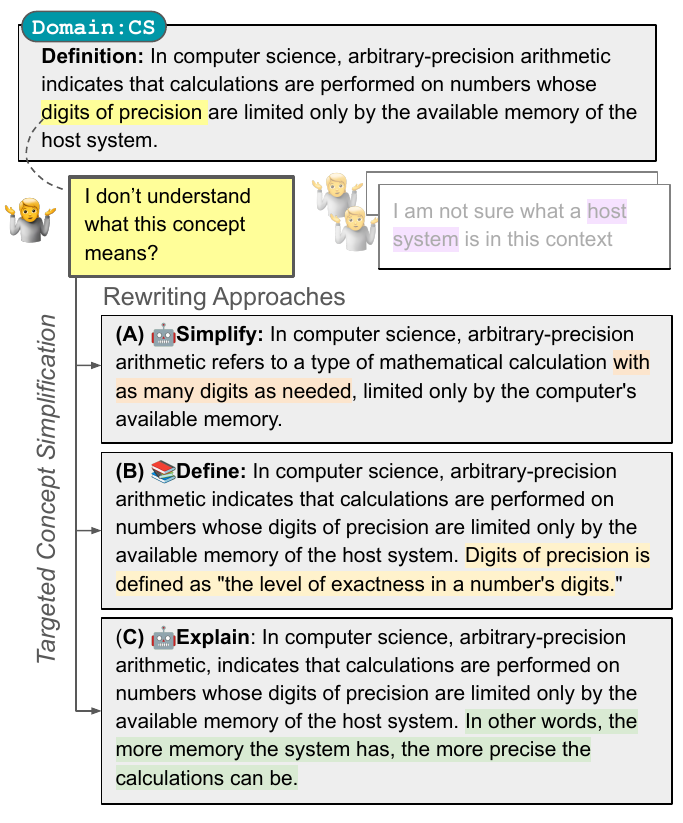}
    \caption{An example from the dataset, which consists of a \definition and a potential \difficultconcept in the text that a reader may struggle with. The task is to rewrite the \definition in a way that simplifies this concept for the reader. 
    (a) Simplifies ``digits of precision'' to ``as many digits as needed'', (b) Adds the definition of ``digits of precision'', (c) Contextually explains that ``digits of precision'' refers to precision of calculations and how it relates to memory.
    }
    \vspace{-0.6cm}
    \label{fig:running_example}
\end{figure}

For example, in Figure~\ref{fig:running_example}, a person unfamiliar with the concept ``digits of precision'' will not understand the definition of ``arbitrary precision arithmetic''. AI tools could help rewrite the text by lexically substituting ``digits of precision'' with ``as many digits as needed'' (simplifying) or by elaborating on the concept (defining or explaining). (a) Lexical simplification makes the definition understandable by reducing the overall complexity, perhaps losing some of the meaning.  (b) Adding a definition of ``digits of precision'' may broaden the reader's vocabulary but does not explain its significance in the context of the overall definition (i.e. its implications for calculations and memory).  (c) Providing a contextual explanation about ``digits of precision'' could more explicitly link the relation between memory and preciseness of calculations, enhancing comprehension~\cite{ van2010using,srikanth-li-2021-elaborative}.

In our study, we asked human raters to read definitions from 13 academic \domain{s} and identify the challenges in understanding them. We found that 50\% of the reading difficulties arose from unfamiliar concepts, and annotators expressed the need for more context around them. Motivated by this, we present the new task of \tasknameemph for rewriting text to support understanding of \difficultconcept{s} within the definitions' context. This task focuses on simplifying specific concepts that users struggle with, allowing for personalized reading support than simply rewriting an entire document at an easier reading level. Personalized support with \difficultconcept{s} can help readers receive more contextually-relevant information tailored to their background knowledge. For instance, a computer scientist reading a physics document might struggle with physics concepts but understand the mathematical terms, while someone without a math background might need help with mathematical terms~\cite{Guo2023PersonalizedJI}.

To investigate this task, we collect a new dataset, \datawiki, consisting of 22k definitions from Wikipedia. We collect definitions using article titles and leading statements from Wikipedia. Our definitions span 13 academic \domain{s} (e.g., business, education, etc., see Table~\ref{tab:wikinumbers}) improving over existing datasets that are limited to a single \domain (e.g., science)~\cite{august-etal-2022-generating}. We annotate a potential \difficultconcept in each definition using an automated heuristic~\cite{biran-etal-2011-putting}.

\begin{table}[t]
    \centering
    \begin{tabular}{p{3.9cm}r}
        \toprule
        \multicolumn{1}{c}{Domain} & \multicolumn{1}{c}{\#Definitions} \\
        \midrule
        Food \& Drink & 1,403\\
        Performing arts & 322\\
        Business \& Economics & 1,539 \\
        Politics \& Government & 2,267\\
        Biology & 7,200\\
        Chemistry & 957\\
        Computing & 2,083\\
        Earth and Environment & 1,314\\
        Mathematics & 1,747\\
        Medicine \& Health & 2,939 \\
        Physics & 741\\
        Engineering &	89\\
        Technology &	7 \\
        \midrule
        Total & 22,561 \\
        \bottomrule
    \end{tabular}
    \caption{Domains and number of definitions in each domain in the \datawiki dataset.}
    \label{tab:wikinumbers}
\end{table}

We use this dataset to evaluate the performance of open-source and commercial LLMs on \tasknameemph. We explore three methods for rewriting definitions: adding a dictionary definition of the \difficultconcept, prompting LLMs to simplify the \difficultconcept, and prompting LLMs to explain the \difficultconcept in context. We conduct human evaluations of all three approaches along three dimensions: 1) meaning preservation, 2) whether a reader who is unfamiliar with the \difficultconcept
can understand the rewritten definition, and 3) whether the rewritten definition is easier to understand than the original.
Our human evaluations demonstrate a clear preference towards strategies for contextual explanation of the \difficultconcept rather than lexical simplifications. However, we also find that LLMs need to improve further on dimensions of comprehension.
Low to mild correlations of automated simplification metrics with human evaluations of comprehension and meaning preservation ($\sim$ 0.1-0.3) also indicate a need for better metrics to evaluate nuanced contextual explanations.

In summary, our main contributions include:

\begin{itemize}
    \item Introducing \taskname as a task for supporting readers as they encounter difficult concepts in text.
    \vspace*{-2.5mm}
    \item Analysis from an annotation study examining the difficulties humans face in reading and the possible utility of assistance in understanding difficult concepts. \vspace*{-2.5mm}
    \item \datawiki, a dataset of 22k challenging domain-specific definitions collected from Wikipedia with automatically-annotated difficult concepts. \vspace*{-2.5mm}
    \item Human evaluations of the performance of open-source and commercial LLMs on our task across multiple quality dimensions, including analysis of different prompting strategies and automatic metrics.
\end{itemize}


\section{Background}
\paragraph{Cognitive Support and Human Reading Comprehension}
Successful reading comprehension is key to integrating new knowledge and fostering learning from text~\cite{lorch1997understanding,dunietz-etal-2020-test}. Cognitive theories suggest that comprehension is a multi-stage process that primarily involves 1) constructing a local meaning representation of text  such as concepts, facts, and their relations~\cite{graesser1994constructing}, and 2) forming a schema and filling in gaps using background knowledge to create a ``mental picture'' of what the text is about~\cite{kintsch1978toward,bartlett1995remembering}. Adult readers lacking domain knowledge can be supported by explicit cues, such as examples and explanations, to help them construct better mental representations of ideas from the text ~\cite{kintsch1991role,van2010using}.

\paragraph{Text Simplification}
Reducing reading-level complexity and syntax~\cite{garbacea-etal-2021-explainable} in text simplification benefits specific audiences like students, second language learners, and individuals with dyslexia~\cite{paetzold-specia-2016-anita,bingel-etal-2018-lexi}, but may not enhance comprehension for general adult readers~\cite{garbacea-etal-2021-explainable}. Contextual explanations can enhance comprehension but findings from studies of elaborating events in news domains~\cite{srikanth-li-2021-elaborative} may not be the same as difficulty with concepts in academic texts.
While Wikipedia and news corpora~\cite{kauchak-2013-improving,xu-etal-2015-problems,zhang-lapata-2017-sentence} have advanced text simplification, they focus more on syntax and discourse difficulties than on academic concepts. Similarly,  lexicons are limited to medicine~\cite{elhadad-sutaria-2007-mining,ong2007simplifying} and science concepts~\cite{august-etal-2022-generating}, highlighting the need for a multi-domain corpus to advance personalized simplification for a general audience.

\paragraph{Complex Terms and Jargon}
Lexical simplification systems~\cite{paetzold-specia-2016-anita} have been shown to benefit children, people with language impairments or medical jargon simplification~\cite{fatima-strube-2023-cross, joseph-etal-2023-multilingual}. However, beyond lab studies, it is challenging to specify reader knowledge in large-scale evaluations. Proxies for audience knowledge include specialized lexicons~\cite{paris-1988-tailoring,elhadad-sutaria-2007-mining},  coarse indicators such as reading grade level~\cite{agrawal-carpuat-2023-controlling}, or binary indicators to denote science knowledge audience~\cite{august-etal-2022-generating}. \citet{Guo2023PersonalizedJI} highlighted the challenge of specifying audience knowledge at a finer level, suggesting the use of \domain as a proxy for concept familiarity. Building on this, we provide a multi-domain corpus of challenging definitions to specify fine-grained audience levels. Unlike prior work on generating definitions~\cite{august-etal-2022-generating} or simplifying all difficult concepts~\cite{fatima-strube-2023-cross}, we focus on rewriting definitions to address specific concept difficulties, enabling readers to leverage their background knowledge and improve comprehension~\cite{kintsch1991role,rello2013}. While previous tools explored simple strategies like adding definitions for complex terms~\cite{bingel-etal-2018-lexi}, we evaluate LLMs that can provide contextual explanations~\cite{srikanth-li-2021-elaborative}.

\section{What AI assistance can benefit reading domain specific text}
\label{sec:us1}

\begin{figure}
    \centering
    \includegraphics[trim={10 25 20 0 },clip,width=1\columnwidth]{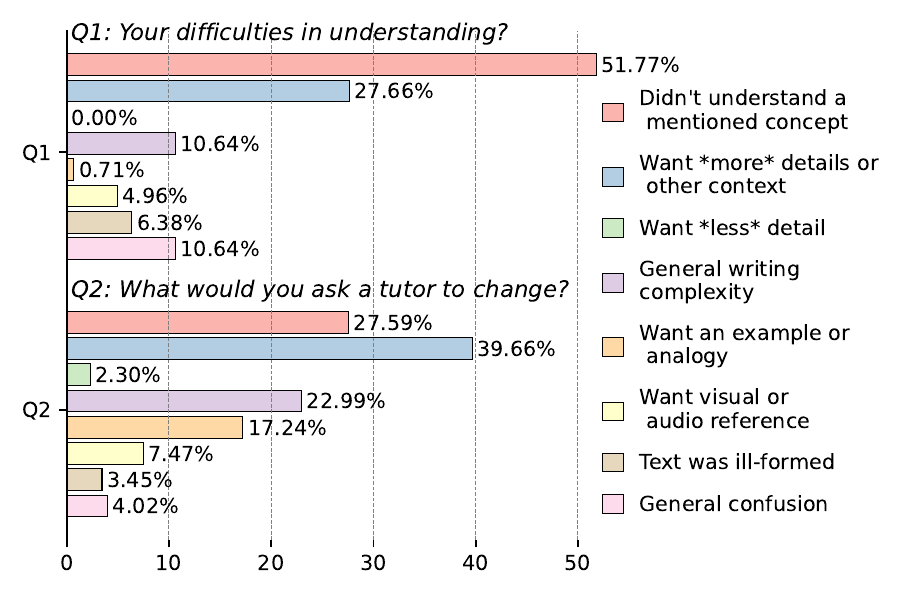}
    \caption{Results of annotator study: We asked annotators to read complex text for (1) what made the text difficult for them to understand and (2) how they would want a tutor to edit the text to help their understanding.}
    \label{fig:us1results}
\end{figure}

To better motivate the scope of this task, we investigate adult readers' difficulties with domain-specific text and what types of help they would want from an AI-tutor. We randomly selected a set of 900 text examples from Wikipedia-derived definitions spanning 13 \domain{s} (see Section~\ref{sec:wikidataset} for details about definitions and domain selection).  For each example, we ask a human annotator to respond in free-text to: 1)~the reasons for difficulty (if any) when reading the definitions, and 2)~what they would ask a tutor to change in the definition if they faced a difficulty. In Figure~\ref{fig:us1results}, we report recurring themes from annotator responses to both questions for cases where annotators had difficulty reading from these examples (categories were agreed on by the authors, see details in Appendix Section~\ref{sec:appendix:userstudy_understanding}).

These results suggest that specific \difficultconcept{\it s} used in the definitions were one of the most frequent reasons for reading difficulty (52\% of definitions had such difficulty), and annotators frequently asked for help from a tutor with these concepts (28\%). This indicates that our proposed task, \taskname{}, is an important subtask for simplification aimed to resolve a key challenge for lay adult readers. When asking a tutor for help, annotators also explicitly asked for \textbf{more} details on the \difficultconcept (rather than less). This suggests that contextual elaborations \citep{srikanth-li-2021-elaborative} for these \difficultconcept{s}  are a better alternative over lexical simplifications to support their comprehension and knowledge. Annotators also asked for examples/analogies (17\%), visual/audio aids (8\%), or identified general issues with the writing complexity (23\%, e.g., an issue with general reading level or syntactic complexity), though to a lesser degree. The majority of our annotators had an above high-school level educational qualification (see Table~\ref{tab:appendix:demographic_bg_user_study1} in Appendix), suggesting that unfamiliar concepts in context is a greater challenge for skilled adult readers than simply difficult words or syntax.


\section{Task Definition}

We present \textit{\taskname{}}: a text simplification task focused on specific words or phrases that readers find difficult to understand. This setup allows for personalized and controlled rewriting of difficult concepts. Our initial user study (Section \ref{sec:us1}) shows that unfamiliar words or phrases often hinder comprehension.

The task of \taskname is to rewrite an input definition containing a concept $c$ to make it understandable to someone unfamiliar with the concept.  For example, Figure~\ref{fig:running_example} shows the definition of the term ``arbitrary precision arithmetic.'' The task is to rewrite the definition to help someone unfamiliar with the \difficultconcept ``digits of precision.''  Possible approaches could involve replacing ``digits of precision'' with a simpler phrase like ``as many digits as needed,'' explaining it within the definition, or perhaps even adding examples, analogies, or illustrations. The usefulness of each strategy will depend on its ability to complement the reader's existing knowledge about the topic~\citet{kintsch1991role}. Unlike other text simplification tasks, our task targets simplifying concepts \textit{difficult for the reader} rather than simplifying the entire text.

\section{The \datawiki Dataset}
\label{sec:wikidataset}
To support research on \taskname, we introduce a dataset of 22k definitions from 13 academic domains\footnote{\href{https://github.com/google-deepmind/wikidomains}{https://github.com/google-deepmind/wikidomains}}, where each \definition{} is a 1--2 sentence explanation of a \concept{}\footnote{We call the concept being explained by the original definition a \concept to avoid confusion with \difficultconcept{\it s} present within the definitions.}.  Within each definition, we select a \difficultconcept---a word or phrase that could impede the reader's ability to comprehend the definition as whole. We take inspiration from \citet{august-etal-2022-generating} who collected definitions from Wikipedia science glossaries; however, instead of glossaries, we directly collect definitions from Wikipedia articles of concepts spanning 13 \textit{\domain{s}} (see Table~\ref{tab:wikinumbers} for list of domains).


To collect definitions, we start with \citet{Johnson2021}'s dataset that contains all English Wikipedia articles with probabilities of belonging to high-level domains. These domains are broad academic topics (e.g., Physics, Economics) that Wikipedia editors identified through consensus~\cite{asthana2018}. 
We refer to each Wikipedia article title as a \concept and take the first sentence of its lead section as its \definition ~\cite{august-etal-2022-generating}. For every domain, we first select articles with domain assignment probabilities greater than a threshold $\delta_{domain}$.\footnote{Manual inspection of topic assignments for 50 Wikipedia articles suggested that a threshold of 0.7 was reasonable to identify articles belonging to a domain.} To filter out low-importance articles that could be named entities, unimportant places or things, we also excluded articles having a page-rank percentile score less than $\delta_{pr}$.\footnote{We determined the threshold as 0.1 through a manual examination of 100 articles.} Finally, we also excluded articles that were additionally members of domains related to named entities, events, or things (e.g., Biography). Table~\ref{tab:wikinumbers} summarizes the 13 domains and the number of articles in each domain that the final \datawiki dataset contains (more details in Appendix~\ref{sec:appendix:dataset}). We also provide each \concept{'s} lead section in the dataset for future research.

We select a training, development, and test split of 15,873/3,384/3,304 examples (see Table~\ref{tab:wikisplitnumbers} for more details about the data splits.) We conduct our experiments in a zero- or few-shot setting without using the training or development data, but we publicly release the full set to facilitate future research.

\begin{table}[]
    \centering
\begin{tabular}{lrrr}
\toprule
 & \multicolumn{1}{c}{train} & \multicolumn{1}{c}{dev} & \multicolumn{1}{c}{test} \\
 \midrule
\# definitions & 15,873 & 3,384 & 3,304 \\
avg \# tokens & 22.75 & 22.63 & 22.61 \\
total \# tokens & 361,066 & 76,572 & 74,691 \\
vocab size & 42,356 & 15,576 & 14,911 \\
 \bottomrule
  \end{tabular}
\caption{Statistics on  \datawiki definitions broken down by split; \#tokens and vocabulary size are calculated by splitting the definitions on whitespace and removing punctuation.}
\label{tab:wikisplitnumbers}
\end{table}

\subsection{Difficult Concept Identification}
\label{sec:difficult_concept}
For each definition, we automatically label a potential \difficultconcept that could impede a reader's comprehension. Lay readers will be more familiar with concepts that are popularly mentioned across Wikipedia (e.g., ``bacteria'') than concepts that only occur in articles of a specific domain (e.g., ``Phytosterol''). Thus, following prior work on approximating word difficulties using specificity-based measures~\cite{biran-etal-2011-putting}, we use a domain-specificity measure to score concept difficulty for a lay audience.


First, we identify candidate concepts $c$ mentioned in each \concept{'s} definition using Wikidata~\cite{vrande2014}\footnote{We used the Wikidata extension in spaCy to identify concepts in definitions that have corresponding Wikidata entries.}. We then order the candidates by a score of how specific they are to the \concept{'s} domain $D_t$. This  is  measured by the ratio of how many articles $\mathcal{A}$ the concept $c$ appears in within this domain compared to across Wikipedia generally:

\begin{equation}
    \label{eq:domainspec}
    \frac{
    \sum_{\mathcal{A} \in \mathcal{D}_t}{\mathbbm{1}[c \in \mathcal{A}]}
    }
    {
    \sum_{\mathcal{A} \in \mathcal{D_\text{all}}}{\mathbbm{1}[c \in \mathcal{A}]}
    }
\end{equation}
For each definition, we select one \difficultconcept out of the top-$k$ identified candidates\footnote{We chose $k$ as 2 based on manual assessment of 100 definitions.}.
If we could not identify any \difficultconcept{} in the definition, we instead chose a difficult concept using the age of acquisition lexicon~\cite{kuperman2012age}, which provides an average age when different words are acquired as a proxy for its difficulty.

\section{Experiments}\label{sec:experiments}
We explore the performance of existing NLP tools on \taskname{} and possible avenues for future improvement. More concretely, we investigate the following research questions:

{\bf RQ1:} What is the performance of out-of-the-box NLP tools in this task?

{\bf RQ2:} Which types of simplification strategies improve human understanding of \difficultconcept{s} and the definitions that they appear in?

{\bf RQ3:} For  \tasknameemph, how do human evaluations compare to automatic metrics commonly used in text simplification?

We perform experiments on the \datawiki test data created in Section~\ref{sec:wikidataset}.  As an additional evaluation set, we also use the scientific definitions dataset from \citet{august-etal-2022-generating} (\scidef) that contains definitions of science terms extracted from Wikipedia glossaries and MedQuAD~\cite{ben2019question}. We perform the same post-processing on \scidef as with \datawiki to select a \difficultconcept{} within each definition.

\begin{table*}[t]
    \centering
    \small
    \begin{tabular}{lp{4cm}p{10.5cm}}
        \toprule
       & Name & Description \\
        \midrule
       \multirow{5}{*}{\rotatebox[origin=c]{90}{\small Human Eval.}}& Meaning preservation (\hmp) & Human evaluation of whether the rewritten definition preserves the meaning of the original definition (on a 5-point Likert scale; 5 = perfectly preserved).\\
        & Rewrite understanding (\hru) & Human evaluation of whether a reader can understand the rewritten definition if they do not know the \difficultconcept (1 = yes, 0 = no).\\
       & Rewrite easier (\hre) & Human evaluation of whether the rewritten definition is easier to understand than the original definition (1~= rewrite is easier; 0 = the original is easier or both are similar).\\
        \midrule

       \multirow{13}{*}{\rotatebox[origin=c]{90}{\small Automatic Eval.}} & Density & Density~\cite{grusky-etal-2018-newsroom} is a measure of how extractive the rewritten definition is from the original definition.\\
       & \bleufour & \bleufour score~\cite{Papineni02bleu:a} of the rewritten definition with respect to the original definition.\\
       & \bertscore (BertSc) & \bertscore~\cite{bert-score} of the rewritten definition with respect to the original definition.\\
       & Change in length ($\Delta$Len) & Average difference between the lengths (in number of tokens) of the rewritten and the original definition (positive means the rewritten definition is longer than the original).\\
      &  Change in age of acquisition ($\Delta$AoA) & Average difference of the top-10 percentile of the age-of-acquisition~\cite{kuperman2012age} of the words between the rewritten and the original definition (positive means the rewritten definition uses less complex words).\\
       & Change in Flesch ease ($\Delta$Flesch) & Average difference of the Flesch reading ease \cite{fleschreadingease} between the rewritten and the original definition (positive means the rewritten definition is at an easier reading level than the original).\\

        \bottomrule
    \end{tabular}
    \caption{Human and automatic metrics used to evaluate LLM rewritten text for concept simplification.}
    \label{tab:metrics}
\end{table*}

\subsection{Models}
To explore the benchmark performance on this data, we selected four popular LLMs: GPT-4 \citep{openai2023gpt4}, PaLM-2 \citep{anil2023palm}, Falcon-40b \citep{falcon40b}, and BLOOM-170b \citep{workshop2023bloom}. For the open-source models, we selected the instruct versions with the highest number of parameters available.

We also included a baseline approach of dictionary look-up (non-LLM) to compare to the LLMs. For this baseline, we looked up a definition of the \difficultconcept{} and simply appended it to the end of the original definition. We retrieved the definition from Wikidata~\cite{vrande2014}, falling back on WordNet \citep{miller-1994-wordnet} if the term was not found in Wikidata (or stated that the \difficultconcept{}'s definition could not be found if both sources failed).

\subsection{Simplification Strategies and Prompts}\label{sec:prompts}
In our preliminary user study (Section~\ref{sec:us1}), we found that users frequently indicated they would like more details and context, as well as more general breakdowns of writing complexity.  These two strategies also correspond to familiar approaches for general text simplification tasks that rely on elaboration \cite{srikanth-li-2021-elaborative} and lexical changes \cite{paetzold-specia-2016-anita}, respectively.

We chose two different prompts for the LLMs that reflect these two simplification strategies.
In our first prompt, we show the model the \concept, \definition, and \difficultconcept.  We instruct the model to rewrite the definition, ``integrating an explanation'' for the difficult concept ($explain$). The second prompt is similar, except we instruct the model to rewrite the definition ``simplifying'' the difficult concept word ($simplify$).

We chose the specific wording of the prompts for the two strategies after a small scale analysis of results with a few candidate prompts. We describe the candidate prompts, and the full phrasing of the final selected prompts in the Appendix (Section~\ref{sec:appendix:prompts}).  We report results using 3-shot settings for the LLMs.\footnote{We also experimented with a zero-shot settings with results in the Appendix.}

\subsection{Human Evaluation}
We asked human raters to rate the rewritten definitions along dimensions of meaning preservation and ease of understanding of the rewrites with respect to both the \difficultconcept{} and the original definition. Specifically, we asked them about (1) {\it meaning preservation}, denoted as \hmp: how much does the rewrite preserve the meaning of the original definition on a Likert scale of 5; (2) {\it rewrite understanding}, denoted as \hru: If a reader is unfamiliar with the \difficultconcept{}, would they be able to understand the rewrite (Yes/No); (3) {\it rewrite easier}, denoted as \hre: Is the rewrite easier to understand than the original? These dimensions are summarized in the first three rows of Table~\ref{tab:metrics}. We obtain judgments  from 3 human raters per example for 120 randomly selected examples from the \datawiki dataset and 60 randomly selected examples from the \scidef dataset (2880 judgments in total). We provide exact wording of the question, their rationale and annotator background in the Appendix (Section~\ref{sec:appendix:userstudy_llm_evals}). In Appendix Table~\ref{tab:appendix:ka} we show Krippendorff's alpha agreement scores for each human evaluation dimension.



\subsection{Automated Metrics}
We investigate the utility of commonly used simplification automated metrics for our task and compare them to human judgments. Because our data are reference-less, we cannot use reference-based metrics like SARI~\cite{xu-etal-2016-optimizing}. Instead, we estimate changes in complexity using the difference between the rewritten and the original definition in terms of: (1) age of acquisition (AoA;~\citealt{kuperman2012age}), (2) Flesch reading ease \cite{fleschreadingease}, and (3) token length. We also measure density \cite{grusky-etal-2018-newsroom}, which scores how extractive the rewritten defintion is from the original. Lastly, we use \bleu~\cite{Papineni02bleu:a} and \bertscore~\cite{bert-score} to score the similarity of the rewritten definitions with respect to the original definition.
Table \ref{tab:metrics} presents a full list of the human and automatic evaluations.

\begin{table}[tb]
    \centering
    \resizebox{0.35\textwidth}{!}{%
    \begin{tabular}{c}
    \begin{tabular}{cllrrr}
    \toprule
         & & {Model}  & {\hmp} & {\hre} & {\hru} \\
    \midrule
    \multirow{9}{*}{\rotatebox[origin=c]{90}{\small \sc WikiDomains}} & & Baseline & {4.66} & 0.31 & {0.79}\\
    \cmidrule{2-6}
    & \multirow{4}{*}{\rotatebox[origin=c]{90}{\small simplify}} & Bloomz & 4.25 & 0.20 & 0.53 \\
    & & Falcon & 3.82 & 0.59 & 0.67 \\
    & & PaLM2 & \textbf{4.71} & 0.12 & 0.46 \\
    & & GPT4 & 4.43 & 0.29 & 0.75 \\
    \cmidrule{2-6}
    & \multirow{4}{*}{\rotatebox[origin=c]{90}{\small explain}} & Bloomz & 4.53 & 0.43 & 0.69 \\
    & & Falcon & 4.30 & 0.55 & 0.71 \\
    & & PaLM2 & 4.64 & 0.59 & 0.66 \\
    & & GPT4 & 4.47 & \textbf{0.75} & \textbf{0.82} \\
    \bottomrule
    \end{tabular} \\

    \newline 
    \begin{tabular}{cllrrr}
    \toprule
         & & {Model}  & {\hmp} & {\hre} & {\hru} \\
    \midrule
    \multirow{9}{*}{\rotatebox[origin=c]{90}{\small \sc SciDef}} & & Baseline & 4.21 & 0.40 & 0.61\\
    \cmidrule{2-6}
    & \multirow{4}{*}{\rotatebox[origin=c]{90}{\small simplify}} & Bloomz & 4.58 & 0.10 & 0.46 \\
    & & Falcon & 4.24 & 0.25 & 0.57 \\
    & & PaLM2 & 4.57 & 0.12 & 0.62 \\
    & & GPT4 & 4.47 & 0.12 & \textbf{0.88} \\
    \cmidrule{2-6}
    & \multirow{4}{*}{\rotatebox[origin=c]{90}{\small explain}} & Bloomz & 4.39 & 0.53 & 0.65 \\
    & & Falcon & 4.29 & 0.53 & 0.86 \\
    & & PaLM2 & \textbf{4.86} & 0.18 & 0.54 \\
    & & GPT4 & 4.09 & \textbf{0.58} & 0.87 \\
    \bottomrule
    \end{tabular}
    \end{tabular}%
    }
    \caption{Human evaluations of LLM-generated rewrites for targeted concept simplification for the metrics \hmp (meaning preservation), \hre (rewrite easier), and \hru (rewrite understanding) in 3-shot setting.}
    \label{tab:h_metrics}
\end{table}

\subsection{Model Rankings}
Table~\ref{tab:h_metrics} summarizes the evaluations of the rewrites based on human judgment according to the meaning preservation (\hmp), whether the rewrite is easier to understand than the original (\hre), and whether the rewrite can be understood for someone unfamiliar with the \difficultconcept (\hru).


We observe that no model excels in all dimensions, though GPT-4 performs best on average. Different models have distinct strengths; for instance, PaLM2 excels in meaning preservation but its rewrites are rarely easier to understand. Additionally, the dictionary-lookup baseline performs comparably well to the LLM models.


Weaker scores on the \hru and \hre dimensions compared to the \hmp dimension across all models, indicates opportunities for future research to improve these scores.

\subsection{Simplifying vs Explaining}
\label{sec:promptresults}
\begin{table}[t]
    \centering

    \begin{tabular}{cllll}
    \toprule
         & {Prompt}& {\hmp}& {\hre} & {\hru}  \\
    \midrule
        \multirow{2}{3.5em}{\small \sc Wiki\\Domains} & simplify &
        4.30&0.30&0.60\\
        & explain &  {\bf 4.48}$^*$ & {\bf 0.57}$^*$ & {\bf 0.72}$^*$ \\
        \midrule
        \multirow{2}{3.5em}{\small \sc SciDef} & simplify & {\bf 4.47} & 0.15 & 0.64 \\
        & explain & 4.41 & {\bf 0.45}$^*$ & {\bf 0.73}$^*$\\
    \bottomrule
    \end{tabular}
    \caption{Comparison of the prompts -- simplify and explain --
    for the human evaluation metrics. All results are significantly different (ttest, $p <0.01$) marked by *, except for the comparison of meaning preservation on \scidef. Results are from the 3-shot setting.}
    \label{tab:prompt_comparison}
\end{table}

We discuss the differences of the human evaluations for the two prompt strategies---\explain and \simplify (introduced in Section~\ref{sec:prompts}). Table~\ref{tab:prompt_comparison} shows the comparison of the prompt strategies on human evaluation dimensions averaged across the four LLMs. Human raters clearly preferred rewrites where the model was asked to explain the \difficultconcept{} in both \hre and \hru judgments. On \datawiki data, human raters also had a significant preference towards the ``explain'' strategy when judging meaning preservation (the difference in \hmp on \scidef was not significant). This aligns with some of our observations from our initial user study (Section~\ref{sec:us1}), which found that humans preferred adding more context (40\%) as opposed to simpler word substitutions (23\%). This highlights that adding elaborative details is very important towards facilitating human understanding centered around difficult concepts. 




\subsection{Correlation between Human and Automated Evaluation}
\begin{figure}[t]
    \centering
    \includegraphics[width=1\linewidth]{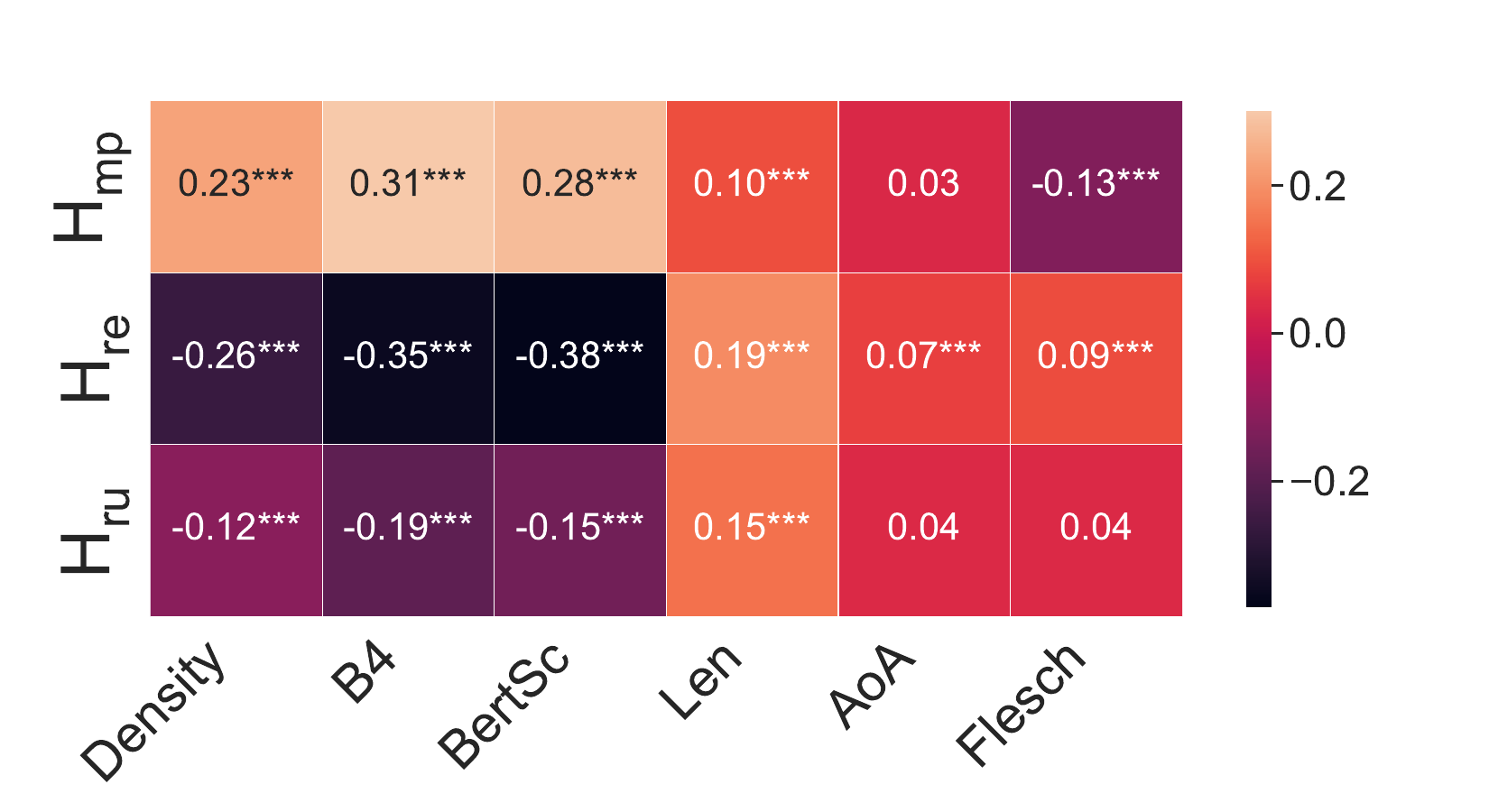}
    \caption{Pearson correlations between automated metrics and human evaluations ($^{***}: p < 0.005, ^{**}: p < 0.05, ^{*}:p < 0.01$).}
    \label{fig:correlations_automated_human}
\end{figure}

Figure~\ref{fig:correlations_automated_human} shows the correlations between automated metrics and human evaluations (\hmp, \hre, \hru). We observe no single metric captures all the dimensions of human evaluations. \bleufour, Density, and \bertscore show mild correlations with \hmp as they capture similarity of text. However, none of automated metrics correlate with either \hre or \hru which capture comprehension related to the target \difficultconcept{}. Even Flesch reading ease, which is commonly used in text simplification setups, is not adequate for measuring whether the rewrites are easier understood. This result calls for new metrics, beyond aggregate similarity measures, to evaluate comprehension at the  semantic level of concepts.

In the Appendix (Table~\ref{tab:automated_metrics}), we show the full automated metric scores for each model, which may be useful in characterizing some qualities of the outputs.  For example, Bloomz and PaLM2 make relatively few changes to the text (low $\Delta$Len), and GPT4 chose considerably easier words in the rewrite (high $\Delta$AoA). GPT4's low meaning preservation rating suggests choosing easier words is not always desirable (Table~\ref{tab:h_metrics}). However, given the low correlations with human scores, we generally keep our observations about relative model rankings to the human judgment.



\begin{table}[t]
    \centering
    \small
    \begin{tabular}{p{7.2cm}}
    \toprule
        (a) Lustre or luster is the way light interacts with the surface of a crystal, rock, or mineral. \DIFaddbegin \DIFadd{Mineral is defined as "naturally occurring usually inorganic substance that has a (more or less) definite chemical composition and a crystal structure."}\DIFaddend \\
        \midrule
        (b) Quality control, or QC for short, is a process by which entities review the quality of all factors involved in production. \DIFaddbegin \DIFadd{Entity is defined as "something that exists in the identified universe."}\DIFaddend. \\
    \bottomrule
    \end{tabular}
    \caption{Examples from the dictionary baseline, which appends a definition shown in \DIFaddbegin \DIFadd{blue}\DIFaddend. (a) Added definition has domain-specific jargon that may be unfamiliar to the reader. (b) Added definition is vague, not accounting for the context.}
    \label{tab:dictionary_baseline_examples}
\end{table}

\begin{table}[t]
    \centering
    \small
    \begin{tabular}{p{7cm}}
    \toprule
        \textbf{Economics}: The Financial Stability Board (FSB) is an international body that monitors and makes recommendations about the \DIFdelbegin \DIFdel{global financial system}\DIFdelend \DIFaddbegin \DIFadd{world's money}\DIFaddend . (\textbf{global financial system}) [PaLM2]\\
        \midrule
        \textbf{Biology}: 
        \DIFdelbegin \DIFdel{Jungle is an area covered with dense vegetation dominated by large trees, often tropical}\DIFdelend \DIFaddbegin \DIFadd{A jungle is a region filled with thick plant life, often dominated by large trees, typically found in tropical areas.}\DIFaddend\ (\textbf{vegetation}) [GPT4]\\
        \midrule
        \textbf{Computing}: Prolog is a logic programming language associated with artificial intelligence and computational linguistics. (\textbf{linguistics})  [Bloomz] [no change]\\
    \bottomrule
    \end{tabular}
    \caption{Examples of concept simplification behavior for the {\it simplify} 3-shot prompt from three domains: Economics, Biology and Computing. The \difficultconcept is shown in \textbf{bold} at the end of the definition. Deletions are show in \DIFdelbegin \DIFdel{red}\DIFdelend; additions in \DIFaddbegin \DIFadd{blue}\DIFaddend.}
    \label{tab:concept_simplify_examples}
    \vspace*{-4mm}
\end{table}

\section{Discussion}
We close our paper by discussing the research questions we posed in our experiments and how they may relate to future improvements on this task.

\paragraph{Can LLMs support Contextual Explanations of Difficult Text?} Despite their instruction-following capabilities, human evaluations indicate that there's still considerable room for improvement at this task. Human judgments (Table~\ref{tab:h_metrics}) reveal that no model excels universally, each having its own strengths and weaknesses. All models tend to perform better at meaning preservation, though other dimensions may be more crucial for enhancing broader comprehension~\cite{kintsch1991role}.

Our evaluations support prior findings that dictionary-based methods for simplification are limited by availability and their inability to personalize to the reader's context and background knowledge~\cite{august-etal-2022-generating}. Table~\ref{tab:dictionary_baseline_examples} shows two examples from the dictionary baseline that are either too vague or too complex to be useful to a lay reader.
%
%
However, we find that LLMs outperform the deterministic dictionary look-up baseline only by a small margin depending on the dimension of quality. In examples of output  (Table~\ref{tab:concept_simplify_examples}), we can see failure cases where the models either over-simplify text beyond just the \difficultconcept{} or make no changes to the definition at all. LLMs have been found to be useful for reading-grade level simplifications~\cite{agrawal-carpuat-2023-controlling}, yet they seem to struggle with making fine-grained simplifications at the level of \difficultconcept{s}, calling for more careful tooling for targeted simplification. While more custom prompts may elicit desired simplifications from LLMs, we cannot expect lay audience to be familiar with such prompting~\cite{pereira2023}. 


\paragraph{Strategies Supporting Readers in Understanding Difficult Text.} Open-ended human feedback about reading difficulties (Section~\ref{sec:us1}) as well as human judgment of differences between prompts (Section~\ref{sec:promptresults}) support the idea that adult readers may prefer additional details and context addition to understand \difficultconcept{s}.  This echoes prior cognitive science work that cues in text (e.g., explanations, examples, analogies) enable readers to effectively utilize their background knowledge for comprehension~\cite{kintsch1991role,van2010using}.

\paragraph{Better Evaluations to Support Text Understanding.} As shown in prior work \cite{manchego2021}, we find that automated metrics cannot capture fine-grained differences in simplification. While there are some correlations between meaning preservation and \bleufour and \bertscore, we did not observe clear correlations of automated metrics with other dimensions of comprehension, such as alleviating difficulty with an unfamiliar concept.  We observe that many of these metrics rely on brittle lexical scoring~\cite{manchego2021}, and it may be necessary for automated metrics to take more of the underlying concept structure of the rewrites into account in order to adequately judge whether the \difficultconcept{} has been explained sufficiently. A separate LLM to score these dimensions more reliably is an option~\cite{wang-etal-2023-chatgpt,chen-etal-2023-exploring-use,Gao2024LLMbasedNE}; however, human judgment currently remains the best standard in this task.

\section{Conclusion}
To support comprehension of domain-specific text for adult readers, we introduced the task of \tasknameemph to study fine-grained simplification of difficult concepts in context. Our human annotation study highlights the importance of aiding users' understanding of these concepts in domain-specific texts. We also introduced \datawiki, a dataset of 22k definitions across 13 academic domains, to support this task. Our findings show a preference for strategies that add explanatory details over simplifying difficult concepts. Human evaluations of LLM rewrites indicate considerable room for improvement, especially for personalized help with difficult concepts. 


\section{Limitations}
Difficulty with concepts varies based on personal knowledge. Thus, it is challenging to build large-scale evaluation corpora for domain-specific concepts. Our dataset of domain-specific concepts is a first step, providing a foundation for future work to study comprehension across domains.

While we used popular LLMs at the time of conducting the human evaluations, we also acknowledge that some of our LLMs may no longer be state-of-the-art when submitting the work. However, we will release our dataset and have described our experimental setup to promote reproducibility of the results with newer LLMs.

We evaluated our work by asking human raters to rate whether they can understand the definitions. However, human reading comprehension is also goal-directed, and different reading goals will evoke different needs for details~\cite{dunietz-etal-2020-test}. The details needed could differ depending on using the text for one's own understanding or using it for communicating it with other people about specific aspects. E.g., a lawyer communicating with engineers about the risks of a technology may need focus on the applications rather than just the understanding of technical concepts. Future work can evaluate how supporting readers with concept simplifications in documents (e.g., explanations, examples, analogies, and illustrations) help them develop a better understanding of the domain in pre-post tests. 


\section{Ethical Considerations}
We extract our domain-specific definitions dataset from Wikipedia, which is publicly available and accessible to all. However, Wikipedia content has a Global North bias because of its editor base, and concepts in our domain-specific dataset will reflect this bias. We also acknowledge the broader educational implications of making definitions easier to understand, and that using LLMs could introduce false information. While in our work we did not observe instances of hallucinations, LLMs may introduce false information when rewriting entire documents or narratives, and we need robust measures to validate the faithfulness of rewritten definition in addressing concept difficulty and providing correct facts. While our evaluations attempt to provide initial insights into LLM's behavior with difficult concepts in domain-specific text, we also acknowledge that concept difficulty is a complex construct, and it can be dependent on a reader's age, educational, and professional background, which future evaluations should consider.

\section*{Acknowledgements}
We thank the anonymous reviewers for their insightful feedback and suggestions. We'd also like to thank Priyanka Agrawal and Slav Petrov for their feedback on the paper manuscript, as well as the other members of the Google DeepMind community for their guidance throughout the project. We also thank the anonymous individuals who participated in our human evaluations and rated the definitions.

\bibliography{anthology.bib, custom.bib}

\begin{thebibliography}{47}
\providecommand{\natexlab}[1]{#1}

\bibitem[{Agrawal and Carpuat(2023)}]{agrawal-carpuat-2023-controlling}
Sweta Agrawal and Marine Carpuat. 2023.
\newblock \href {https://doi.org/10.18653/v1/2023.emnlp-main.790} {Controlling
  pre-trained language models for grade-specific text simplification}.
\newblock In \emph{Proceedings of the 2023 Conference on Empirical Methods in
  Natural Language Processing}, pages 12807--12819, Singapore. Association for
  Computational Linguistics.

\bibitem[{Almazrouei et~al.(2023)Almazrouei, Alobeidli, Alshamsi, Cappelli,
  Cojocaru, Debbah, Goffinet, Heslow, Launay, Malartic, Noune, Pannier, and
  Penedo}]{falcon40b}
Ebtesam Almazrouei, Hamza Alobeidli, Abdulaziz Alshamsi, Alessandro Cappelli,
  Ruxandra Cojocaru, Merouane Debbah, Etienne Goffinet, Daniel Heslow, Julien
  Launay, Quentin Malartic, Badreddine Noune, Baptiste Pannier, and Guilherme
  Penedo. 2023.
\newblock {Falcon-40B}: an open large language model with state-of-the-art
  performance.

\bibitem[{Alva-Manchego et~al.(2021)Alva-Manchego, Scarton, and
  Specia}]{manchego2021}
Fernando Alva-Manchego, Carolina Scarton, and Lucia Specia. 2021.
\newblock \href {https://doi.org/10.1162/coli_a_00418} {{The (Un)Suitability of
  Automatic Evaluation Metrics for Text Simplification}}.
\newblock \emph{Computational Linguistics}, 47(4):861--889.

\bibitem[{Anil et~al.(2023)Anil, Dai, Firat, Johnson, Lepikhin, Passos,
  Shakeri, Taropa, Bailey, Chen, Chu, Clark, Shafey, Huang, Meier-Hellstern,
  Mishra, Moreira, Omernick, Robinson, Ruder, Tay, Xiao, Xu, Zhang, Abrego,
  Ahn, Austin, Barham, Botha, Bradbury, Brahma, Brooks, Catasta, Cheng, Cherry,
  Choquette-Choo, Chowdhery, Crepy, Dave, Dehghani, Dev, Devlin, Díaz, Du,
  Dyer, Feinberg, Feng, Fienber, Freitag, Garcia, Gehrmann, Gonzalez, Gur-Ari,
  Hand, Hashemi, Hou, Howland, Hu, Hui, Hurwitz, Isard, Ittycheriah, Jagielski,
  Jia, Kenealy, Krikun, Kudugunta, Lan, Lee, Lee, Li, Li, Li, Li, Li, Lim, Lin,
  Liu, Liu, Maggioni, Mahendru, Maynez, Misra, Moussalem, Nado, Nham, Ni,
  Nystrom, Parrish, Pellat, Polacek, Polozov, Pope, Qiao, Reif, Richter, Riley,
  Ros, Roy, Saeta, Samuel, Shelby, Slone, Smilkov, So, Sohn, Tokumine, Valter,
  Vasudevan, Vodrahalli, Wang, Wang, Wang, Wang, Wieting, Wu, Xu, Xu, Xue, Yin,
  Yu, Zhang, Zheng, Zheng, Zhou, Zhou, Petrov, and Wu}]{anil2023palm}
Rohan Anil, Andrew~M. Dai, Orhan Firat, Melvin Johnson, Dmitry Lepikhin,
  Alexandre Passos, Siamak Shakeri, Emanuel Taropa, Paige Bailey, Zhifeng Chen,
  Eric Chu, Jonathan~H. Clark, Laurent~El Shafey, Yanping Huang, Kathy
  Meier-Hellstern, Gaurav Mishra, Erica Moreira, Mark Omernick, Kevin Robinson,
  Sebastian Ruder, Yi~Tay, Kefan Xiao, Yuanzhong Xu, Yujing Zhang,
  Gustavo~Hernandez Abrego, Junwhan Ahn, Jacob Austin, Paul Barham, Jan Botha,
  James Bradbury, Siddhartha Brahma, Kevin Brooks, Michele Catasta, Yong Cheng,
  Colin Cherry, Christopher~A. Choquette-Choo, Aakanksha Chowdhery, Clément
  Crepy, Shachi Dave, Mostafa Dehghani, Sunipa Dev, Jacob Devlin, Mark Díaz,
  Nan Du, Ethan Dyer, Vlad Feinberg, Fangxiaoyu Feng, Vlad Fienber, Markus
  Freitag, Xavier Garcia, Sebastian Gehrmann, Lucas Gonzalez, Guy Gur-Ari,
  Steven Hand, Hadi Hashemi, Le~Hou, Joshua Howland, Andrea Hu, Jeffrey Hui,
  Jeremy Hurwitz, Michael Isard, Abe Ittycheriah, Matthew Jagielski, Wenhao
  Jia, Kathleen Kenealy, Maxim Krikun, Sneha Kudugunta, Chang Lan, Katherine
  Lee, Benjamin Lee, Eric Li, Music Li, Wei Li, YaGuang Li, Jian Li, Hyeontaek
  Lim, Hanzhao Lin, Zhongtao Liu, Frederick Liu, Marcello Maggioni, Aroma
  Mahendru, Joshua Maynez, Vedant Misra, Maysam Moussalem, Zachary Nado, John
  Nham, Eric Ni, Andrew Nystrom, Alicia Parrish, Marie Pellat, Martin Polacek,
  Alex Polozov, Reiner Pope, Siyuan Qiao, Emily Reif, Bryan Richter, Parker
  Riley, Alex~Castro Ros, Aurko Roy, Brennan Saeta, Rajkumar Samuel, Renee
  Shelby, Ambrose Slone, Daniel Smilkov, David~R. So, Daniel Sohn, Simon
  Tokumine, Dasha Valter, Vijay Vasudevan, Kiran Vodrahalli, Xuezhi Wang,
  Pidong Wang, Zirui Wang, Tao Wang, John Wieting, Yuhuai Wu, Kelvin Xu, Yunhan
  Xu, Linting Xue, Pengcheng Yin, Jiahui Yu, Qiao Zhang, Steven Zheng,
  Ce~Zheng, Weikang Zhou, Denny Zhou, Slav Petrov, and Yonghui Wu. 2023.
\newblock \href {https://arxiv.org/abs/2305.10403} {Palm 2 technical report}.
\newblock \emph{Preprint}, arXiv:2305.10403.

\bibitem[{Asthana and Halfaker(2018)}]{asthana2018}
Sumit Asthana and Aaron Halfaker. 2018.
\newblock \href {https://doi.org/10.1145/3274290} {With few eyes, all hoaxes
  are deep}.
\newblock \emph{Proc. ACM Hum.-Comput. Interact.}, 2(CSCW).

\bibitem[{August et~al.(2022)August, Reinecke, and
  Smith}]{august-etal-2022-generating}
Tal August, Katharina Reinecke, and Noah~A. Smith. 2022.
\newblock \href {https://doi.org/10.18653/v1/2022.acl-long.569} {Generating
  scientific definitions with controllable complexity}.
\newblock In \emph{Proceedings of the 60th Annual Meeting of the Association
  for Computational Linguistics (Volume 1: Long Papers)}, pages 8298--8317,
  Dublin, Ireland. Association for Computational Linguistics.

\bibitem[{Bartlett(1995)}]{bartlett1995remembering}
Frederic~Charles Bartlett. 1995.
\newblock \emph{Remembering: A study in experimental and social psychology}.
\newblock Cambridge university press.

\bibitem[{Ben~Abacha and Demner-Fushman(2019)}]{ben2019question}
Asma Ben~Abacha and Dina Demner-Fushman. 2019.
\newblock A question-entailment approach to question answering.
\newblock \emph{BMC bioinformatics}, 20(1):1--23.

\bibitem[{{BigScience Workshop}(2023)}]{workshop2023bloom}
{BigScience Workshop}. 2023.
\newblock \href {https://arxiv.org/abs/2211.05100} {Bloom: A 176b-parameter
  open-access multilingual language model}.
\newblock \emph{Preprint}, arXiv:2211.05100.

\bibitem[{Bingel et~al.(2018)Bingel, Paetzold, and
  S{\o}gaard}]{bingel-etal-2018-lexi}
Joachim Bingel, Gustavo Paetzold, and Anders S{\o}gaard. 2018.
\newblock \href {https://aclanthology.org/C18-1021} {{L}exi: A tool for
  adaptive, personalized text simplification}.
\newblock In \emph{Proceedings of the 27th International Conference on
  Computational Linguistics}, pages 245--258, Santa Fe, New Mexico, USA.
  Association for Computational Linguistics.

\bibitem[{Biran et~al.(2011)Biran, Brody, and
  Elhadad}]{biran-etal-2011-putting}
Or~Biran, Samuel Brody, and No{\'e}mie Elhadad. 2011.
\newblock \href {https://aclanthology.org/P11-2087} {Putting it simply: a
  context-aware approach to lexical simplification}.
\newblock In \emph{Proceedings of the 49th Annual Meeting of the Association
  for Computational Linguistics: Human Language Technologies}, pages 496--501,
  Portland, Oregon, USA. Association for Computational Linguistics.

\bibitem[{Chen et~al.(2023)Chen, Wang, Jiang, Shi, and
  Xu}]{chen-etal-2023-exploring-use}
Yi~Chen, Rui Wang, Haiyun Jiang, Shuming Shi, and Ruifeng Xu. 2023.
\newblock \href {https://doi.org/10.18653/v1/2023.findings-ijcnlp.32}
  {Exploring the use of large language models for reference-free text quality
  evaluation: An empirical study}.
\newblock In \emph{Findings of the Association for Computational Linguistics:
  IJCNLP-AACL 2023 (Findings)}, pages 361--374, Nusa Dua, Bali. Association for
  Computational Linguistics.

\bibitem[{Dettmers et~al.(2022)Dettmers, Lewis, Belkada, and
  Zettlemoyer}]{dettmers2022llm}
Tim Dettmers, Mike Lewis, Younes Belkada, and Luke Zettlemoyer. 2022.
\newblock Llm. int8 (): 8-bit matrix multiplication for transformers at scale.
\newblock \emph{arXiv preprint arXiv:2208.07339}.

\bibitem[{Dunietz et~al.(2020)Dunietz, Burnham, Bharadwaj, Rambow, Chu-Carroll,
  and Ferrucci}]{dunietz-etal-2020-test}
Jesse Dunietz, Greg Burnham, Akash Bharadwaj, Owen Rambow, Jennifer
  Chu-Carroll, and Dave Ferrucci. 2020.
\newblock \href {https://doi.org/10.18653/v1/2020.acl-main.701} {To test
  machine comprehension, start by defining comprehension}.
\newblock In \emph{Proceedings of the 58th Annual Meeting of the Association
  for Computational Linguistics}, pages 7839--7859, Online. Association for
  Computational Linguistics.

\bibitem[{Elhadad and Sutaria(2007)}]{elhadad-sutaria-2007-mining}
Noemie Elhadad and Komal Sutaria. 2007.
\newblock \href {https://aclanthology.org/W07-1007} {Mining a lexicon of
  technical terms and lay equivalents}.
\newblock In \emph{Biological, translational, and clinical language
  processing}, pages 49--56, Prague, Czech Republic. Association for
  Computational Linguistics.

\bibitem[{Fatima and Strube(2023)}]{fatima-strube-2023-cross}
Mehwish Fatima and Michael Strube. 2023.
\newblock \href {https://doi.org/10.18653/v1/2023.acl-long.103} {Cross-lingual
  science journalism: Select, simplify and rewrite summaries for non-expert
  readers}.
\newblock In \emph{Proceedings of the 61st Annual Meeting of the Association
  for Computational Linguistics (Volume 1: Long Papers)}, pages 1843--1861,
  Toronto, Canada. Association for Computational Linguistics.

\bibitem[{Flesch(1948)}]{fleschreadingease}
Rudolf~Franz Flesch. 1948.
\newblock \href {https://api.semanticscholar.org/CorpusID:39344661} {A new
  readability yardstick.}
\newblock \emph{The Journal of applied psychology}, 32 3:221--33.

\bibitem[{Gao et~al.(2024)Gao, Hu, Ruan, Pu, and Wan}]{Gao2024LLMbasedNE}
Mingqi Gao, Xinyu Hu, Jie Ruan, Xiao Pu, and Xiaojun Wan. 2024.
\newblock \href {https://api.semanticscholar.org/CorpusID:267406182} {Llm-based
  nlg evaluation: Current status and challenges}.
\newblock \emph{ArXiv}, abs/2402.01383.

\bibitem[{Garbacea et~al.(2021)Garbacea, Guo, Carton, and
  Mei}]{garbacea-etal-2021-explainable}
Cristina Garbacea, Mengtian Guo, Samuel Carton, and Qiaozhu Mei. 2021.
\newblock \href {https://doi.org/10.18653/v1/2021.acl-long.88} {Explainable
  prediction of text complexity: The missing preliminaries for text
  simplification}.
\newblock In \emph{Proceedings of the 59th Annual Meeting of the Association
  for Computational Linguistics and the 11th International Joint Conference on
  Natural Language Processing (Volume 1: Long Papers)}, pages 1086--1097,
  Online. Association for Computational Linguistics.

\bibitem[{Graesser et~al.(1994)Graesser, Singer, and
  Trabasso}]{graesser1994constructing}
Arthur~C Graesser, Murray Singer, and Tom Trabasso. 1994.
\newblock Constructing inferences during narrative text comprehension.
\newblock \emph{Psychological review}, 101(3):371.

\bibitem[{Grusky et~al.(2018)Grusky, Naaman, and
  Artzi}]{grusky-etal-2018-newsroom}
Max Grusky, Mor Naaman, and Yoav Artzi. 2018.
\newblock \href {https://doi.org/10.18653/v1/N18-1065} {{N}ewsroom: A dataset
  of 1.3 million summaries with diverse extractive strategies}.
\newblock In \emph{Proceedings of the 2018 Conference of the North {A}merican
  Chapter of the Association for Computational Linguistics: Human Language
  Technologies, Volume 1 (Long Papers)}, pages 708--719, New Orleans,
  Louisiana. Association for Computational Linguistics.

\bibitem[{Guo et~al.(2023)Guo, Chang, Antoniak, Bransom, Cohen, Wang, and
  August}]{Guo2023PersonalizedJI}
Yue Guo, Joseph~Chee Chang, Maria Antoniak, Erin Bransom, Trevor Cohen, Lucy~Lu
  Wang, and Tal August. 2023.
\newblock \href {https://api.semanticscholar.org/CorpusID:265220712}
  {Personalized jargon identification for enhanced interdisciplinary
  communication}.
\newblock \emph{ArXiv}, abs/2311.09481.

\bibitem[{Johnson(2021)}]{Johnson2021}
Isaac Johnson. 2021.
\newblock \href {https://doi.org/10.6084/m9.figshare.12619766.v3} {{Wikipedia
  Article Topics for All Languages (based on article outlinks)}}.

\bibitem[{Joseph et~al.(2023)Joseph, Kazanas, Reina, Ramanathan, Xu, Wallace,
  and Li}]{joseph-etal-2023-multilingual}
Sebastian Joseph, Kathryn Kazanas, Keziah Reina, Vishnesh Ramanathan, Wei Xu,
  Byron Wallace, and Junyi~Jessy Li. 2023.
\newblock \href {https://doi.org/10.18653/v1/2023.emnlp-main.1037}
  {Multilingual simplification of medical texts}.
\newblock In \emph{Proceedings of the 2023 Conference on Empirical Methods in
  Natural Language Processing}, pages 16662--16692, Singapore. Association for
  Computational Linguistics.

\bibitem[{Kauchak(2013)}]{kauchak-2013-improving}
David Kauchak. 2013.
\newblock \href {https://aclanthology.org/P13-1151} {Improving text
  simplification language modeling using unsimplified text data}.
\newblock In \emph{Proceedings of the 51st Annual Meeting of the Association
  for Computational Linguistics (Volume 1: Long Papers)}, pages 1537--1546,
  Sofia, Bulgaria. Association for Computational Linguistics.

\bibitem[{Kew et~al.(2023)Kew, Chi, V{\'a}squez-Rodr{\'\i}guez, Agrawal,
  Aumiller, Alva-Manchego, and Shardlow}]{kew-etal-2023-bless}
Tannon Kew, Alison Chi, Laura V{\'a}squez-Rodr{\'\i}guez, Sweta Agrawal, Dennis
  Aumiller, Fernando Alva-Manchego, and Matthew Shardlow. 2023.
\newblock \href {https://doi.org/10.18653/v1/2023.emnlp-main.821} {{BLESS}:
  Benchmarking large language models on sentence simplification}.
\newblock In \emph{Proceedings of the 2023 Conference on Empirical Methods in
  Natural Language Processing}, pages 13291--13309, Singapore. Association for
  Computational Linguistics.

\bibitem[{Kintsch(1991)}]{kintsch1991role}
Walter Kintsch. 1991.
\newblock The role of knowledge in discourse comprehension: A
  construction-integration model.
\newblock In \emph{Advances in psychology}, volume~79, pages 107--153.
  Elsevier.

\bibitem[{Kintsch and Van~Dijk(1978)}]{kintsch1978toward}
Walter Kintsch and Teun~A Van~Dijk. 1978.
\newblock Toward a model of text comprehension and production.
\newblock \emph{Psychological review}, 85(5):363.

\bibitem[{Kuperman et~al.(2012)Kuperman, Stadthagen-Gonzalez, and
  Brysbaert}]{kuperman2012age}
Victor Kuperman, Hans Stadthagen-Gonzalez, and Marc Brysbaert. 2012.
\newblock Age-of-acquisition ratings for 30,000 english words.
\newblock \emph{Behavior research methods}, 44:978--990.

\bibitem[{Lorch~Jr and van~den Broek(1997)}]{lorch1997understanding}
Robert~F Lorch~Jr and Paul van~den Broek. 1997.
\newblock Understanding reading comprehension: Current and future contributions
  of cognitive science.
\newblock \emph{Contemporary educational psychology}, 22(2):213--246.

\bibitem[{Miller(1994)}]{miller-1994-wordnet}
George~A. Miller. 1994.
\newblock \href {https://aclanthology.org/H94-1111} {{W}ord{N}et: A lexical
  database for {E}nglish}.
\newblock In \emph{{H}uman {L}anguage {T}echnology: Proceedings of a Workshop
  held at {P}lainsboro, {N}ew {J}ersey, {M}arch 8-11, 1994}.

\bibitem[{Nielsen et~al.(2002)Nielsen, Clemmensen, and
  Yssing}]{nielsen2002getting}
Janni Nielsen, Torkil Clemmensen, and Carsten Yssing. 2002.
\newblock Getting access to what goes on in people's heads? reflections on the
  think-aloud technique.
\newblock In \emph{Proceedings of the second Nordic conference on
  Human-computer interaction}, pages 101--110.

\bibitem[{Ong et~al.(2007)Ong, Damay, Lojico, Lu, and
  Tarantan}]{ong2007simplifying}
Ethel Ong, Jerwin Damay, Gerard Lojico, Kimberly Lu, and Dex Tarantan. 2007.
\newblock Simplifying text in medical literature.
\newblock \emph{Journal of Research in Science, Computing and Engineering},
  4(1):37--47.

\bibitem[{Open{AI}(2023)}]{openai2023gpt4}
Open{AI}. 2023.
\newblock \href {https://arxiv.org/abs/2303.08774} {Gpt-4 technical report}.
\newblock \emph{Preprint}, arXiv:2303.08774.

\bibitem[{Paetzold and Specia(2016)}]{paetzold-specia-2016-anita}
Gustavo Paetzold and Lucia Specia. 2016.
\newblock \href {https://aclanthology.org/C16-2017} {{A}nita: An intelligent
  text adaptation tool}.
\newblock In \emph{Proceedings of {COLING} 2016, the 26th International
  Conference on Computational Linguistics: System Demonstrations}, pages
  79--83, Osaka, Japan. The COLING 2016 Organizing Committee.

\bibitem[{Papineni et~al.(2002)Papineni, Roukos, Ward, and jing
  Zhu}]{Papineni02bleu:a}
Kishore Papineni, Salim Roukos, Todd Ward, and Wei jing Zhu. 2002.
\newblock Bleu: a method for automatic evaluation of machine translation.
\newblock pages 311--318.

\bibitem[{Paris(1988)}]{paris-1988-tailoring}
Cecile~L. Paris. 1988.
\newblock \href {https://aclanthology.org/J88-3006} {Tailoring object
  descriptions to a user{'}s level of expertise}.
\newblock \emph{Computational Linguistics}, 14(3):64--78.

\bibitem[{Rello et~al.(2013)Rello, Baeza-Yates, Bott, and Saggion}]{rello2013}
Luz Rello, Ricardo Baeza-Yates, Stefan Bott, and Horacio Saggion. 2013.
\newblock \href {https://doi.org/10.1145/2461121.2461126} {Simplify or help?
  text simplification strategies for people with dyslexia}.
\newblock In \emph{Proceedings of the 10th International Cross-Disciplinary
  Conference on Web Accessibility}, W4A '13, New York, NY, USA. Association for
  Computing Machinery.

\bibitem[{Srikanth and Li(2021)}]{srikanth-li-2021-elaborative}
Neha Srikanth and Junyi~Jessy Li. 2021.
\newblock \href {https://doi.org/10.18653/v1/2021.findings-acl.455}
  {Elaborative simplification: Content addition and explanation generation in
  text simplification}.
\newblock In \emph{Findings of the Association for Computational Linguistics:
  ACL-IJCNLP 2021}, pages 5123--5137, Online. Association for Computational
  Linguistics.

\bibitem[{Van~den Broek(2010)}]{van2010using}
Paul Van~den Broek. 2010.
\newblock Using texts in science education: Cognitive processes and knowledge
  representation.
\newblock \emph{Science}, 328(5977):453--456.

\bibitem[{Vrande\v{c}i\'{c} and Kr\"{o}tzsch(2014)}]{vrande2014}
Denny Vrande\v{c}i\'{c} and Markus Kr\"{o}tzsch. 2014.
\newblock \href {https://doi.org/10.1145/2629489} {Wikidata: a free
  collaborative knowledgebase}.
\newblock \emph{Commun. ACM}, 57(10):78–85.

\bibitem[{Wang et~al.(2023)Wang, Liang, Meng, Sun, Shi, Li, Xu, Qu, and
  Zhou}]{wang-etal-2023-chatgpt}
Jiaan Wang, Yunlong Liang, Fandong Meng, Zengkui Sun, Haoxiang Shi, Zhixu Li,
  Jinan Xu, Jianfeng Qu, and Jie Zhou. 2023.
\newblock \href {https://doi.org/10.18653/v1/2023.newsum-1.1} {Is {C}hat{GPT} a
  good {NLG} evaluator? a preliminary study}.
\newblock In \emph{Proceedings of the 4th New Frontiers in Summarization
  Workshop}, pages 1--11, Singapore. Association for Computational Linguistics.

\bibitem[{Xu et~al.(2015)Xu, Callison-Burch, and
  Napoles}]{xu-etal-2015-problems}
Wei Xu, Chris Callison-Burch, and Courtney Napoles. 2015.
\newblock \href {https://doi.org/10.1162/tacl_a_00139} {Problems in current
  text simplification research: New data can help}.
\newblock \emph{Transactions of the Association for Computational Linguistics},
  3:283--297.

\bibitem[{Xu et~al.(2016)Xu, Napoles, Pavlick, Chen, and
  Callison-Burch}]{xu-etal-2016-optimizing}
Wei Xu, Courtney Napoles, Ellie Pavlick, Quanze Chen, and Chris Callison-Burch.
  2016.
\newblock \href {https://www.aclweb.org/anthology/Q16-1029} {Optimizing
  statistical machine translation for text simplification}.
\newblock volume~4, pages 401--415.

\bibitem[{Zamfirescu-Pereira et~al.(2023)Zamfirescu-Pereira, Wong, Hartmann,
  and Yang}]{pereira2023}
J.D. Zamfirescu-Pereira, Richmond~Y. Wong, Bjoern Hartmann, and Qian Yang.
  2023.
\newblock \href {https://doi.org/10.1145/3544548.3581388} {Why johnny can’t
  prompt: How non-ai experts try (and fail) to design llm prompts}.
\newblock In \emph{Proceedings of the 2023 CHI Conference on Human Factors in
  Computing Systems}, CHI '23, New York, NY, USA. Association for Computing
  Machinery.

\bibitem[{Zhang* et~al.(2020)Zhang*, Kishore*, Wu*, Weinberger, and
  Artzi}]{bert-score}
Tianyi Zhang*, Varsha Kishore*, Felix Wu*, Kilian~Q. Weinberger, and Yoav
  Artzi. 2020.
\newblock \href {https://openreview.net/forum?id=SkeHuCVFDr} {Bertscore:
  Evaluating text generation with bert}.
\newblock In \emph{International Conference on Learning Representations}.

\bibitem[{Zhang and Lapata(2017)}]{zhang-lapata-2017-sentence}
Xingxing Zhang and Mirella Lapata. 2017.
\newblock \href {https://doi.org/10.18653/v1/D17-1062} {Sentence simplification
  with deep reinforcement learning}.
\newblock In \emph{Proceedings of the 2017 Conference on Empirical Methods in
  Natural Language Processing}, pages 584--594, Copenhagen, Denmark.
  Association for Computational Linguistics.

\end{thebibliography}
\appendix

\section{User study for understanding difficulty with definitions}
\label{sec:appendix:userstudy_understanding}
As a preliminary study of reading difficulty, we asked annotators to read 900 concept definitions from \datawiki and describe difficulties they have in understanding the definition text.

\begin{figure}
    \centering
    \includegraphics[width=\linewidth]{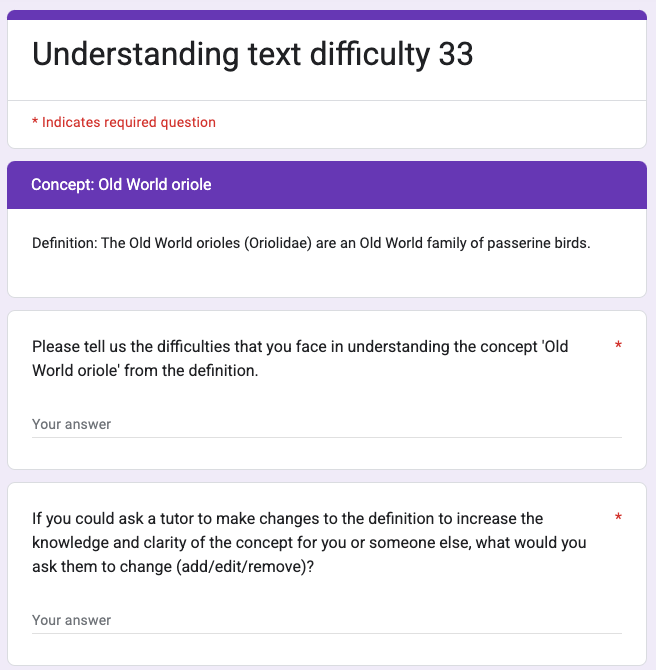}
    \caption{Screenshot of an annotation example for understanding difficulties that readers face with \domain specific text.}
    \label{fig:study_understanding_screenshot}
\end{figure}

As shown in Figure~\ref{fig:study_understanding_screenshot}
we asked each participant the following questions 1) ``Please tell us the difficulties that you face in understanding the concept C from the definition,'' 2) ``If you could ask a tutor to make changes to the definition to increase the knowledge and clarity of the concept for you or someone else, what would you ask them to change (add/edit/remove).'' The first question attempts to understand the difficulties that lay people may face with \domain specific definitions. The second question attempts to involve users in the thinking process of asking a tutor to rewrite the definition. Prior studies in human-centered research suggest that involving users in the task elicits better task-specific challenges than simply asking about the difficulties alone~\cite{nielsen2002getting}.  We keep the task open-ended and ask for free-text responses to give annotators freedom to express any challenges in reading the material.

Following the completion of the study, two of the authors reviewed a random subset of 900 responses (450 responses to Q1 and 450 responses to Q2).  They agreed that when annotators had issues with the reading material, it could generally fit into categories below:
\begin{itemize}
\item Didn't understand a mentioned concept: The annotator referenced a specific word or phrase that was mentioned in the text that hindered their understanding
\item Want *more* details or other context:  The annotator referenced missing details or additional background context that would have helped their understanding
\item Want *less* detail:  the annotator said that the definition text included unnecessary detail
\item General writing complexity: the annotator referenced the overall reading level, syntactic, or lexical complexity of the text
\item Want an example or analogy: the annotator said that an example or analogy would be needed for their understanding
\item Want visual or audio reference: that annotator said that they needed visual or auditory supplements for understanding the text
\item Text was ill-formed: the annotator said the text was ill-formed in some way that made it difficult to read
\item General confusion: annotator expressed general confusion without listing a specific pain point
\end{itemize}

For the responses where an annotator identified a difficulty (which is 57\% of the responses), each was labelled with one or more of the categories above (i.e., categories are not mutually exclusive).  The results of this grouping is summarized in Figure~\ref{fig:us1results}.

\subsection{User demographics for evaluations of difficulty with definitions}
\label{sec:appendix:user_difficulty_demographics}
Table~\ref{tab:appendix:demographic_bg_user_study1} summarizes the demographics of participants who evaluated the difficulty with domain-specific definitions.

\begin{table}[h]
    \centering
    \begin{tabular}{lc}
    \toprule
        Background & Percentage \\
    \midrule
        4-year college degree &               53\%\\
        Master's degree &                      12\%\\
        2-year college degree &                18\%\\
        Some college &                         10\%\\
        High school &                           3\%\\
        Professional degree (MD, JD, etc) &     2\%\\
        Doctoral degree (PhD) &                 2\%\\
    \bottomrule
    \end{tabular}
    \caption{Educational background of annotators for human evaluations of difficulty with definitions. Total number of annotators was 28.}
    \label{tab:appendix:demographic_bg_user_study1}
\end{table}

\section{\datawiki dataset construction}
\label{sec:appendix:dataset}

Editors on Wikipedia have voluntarily come together to form focus groups, called WikiProjects, dedicated to curating and improving articles in specific \domain{s} or interest areas, such as Economics, Chemistry, Literature~\cite{asthana2018}. Any Wikipedia editor can join different WikiProjects and participate in editing articles in that specific WikiProject. As part of the WikiProject effort, Wikipedia editors have annotated a large number of articles on Wikipedia with their WikiProject topic assignments and developed a hierarchical taxonomy of topics called the Wikiprojects directory\footnote{\url{https://en.wikipedia.org/wiki/Wikipedia:WikiProject_Directory}}.
The dataset contains articles from the entire Wikipedia annotated by \domain{s} (broad academic topics) derived from Wikiprojects. We use the topics in the first two levels of this categorization as \domain{s} because they represent broad \domain categorizations.

\subsection{Domain selection criteria}
We selected domains where majority of articles related to names of academic concepts or processes in the domain. Thus, we needed to exclude articles about people, events, names of things (e.g., music albums). For example, the Biography domain contains biographical articles of famous personalities, and the Military domain contains articles on historical military conflicts. To identify such domains, the lead author manually assessed a random sample of 100 articles in each domain. If the number of articles in each domain that corresponded to named entities exceeded 50\% of the assessed articles, we dropped that domain. This is because our work is focused on academically challenging concepts and how they are explained in terms of other concepts. While articles of named entities may contain challenging concepts, the concepts themselves and their explanations in the domain is not the focus of the article. E.g., an article on World War II will likely contain concepts like ``diplomacy'', but its explanation will not be the main focus of the article. We finally excluded the domains: Internet-culture, Literature, Religion, History, Geography, Military-and-warfare, Transportation, Society, Sports, Libraries and Information, Space, and STEM.STEM* (because this is a superset of the domains: Physics, Chemistry, Mathematics, Biology).



\subsection{Dataset snapshot}
\label{sec:appendix:datasetsnapshot}
Table~\ref{tab:appendix:wikidomains_snapshot} shows a snapshot of the \datawiki dataset.

\begin{table*}[h]
    \centering
    \begin{tabular}{p{3cm}p{1cm}p{2.5cm}p{6cm}p{1.5cm}}
    \toprule
        Term & Topic & Wikipedia lead section  & Definition & Difficult concept \\
    \midrule
        Electron gun & Physics &An electron gun...by the number of electrodes.& An electron gun (also called electron emitter) is an electrical component in some vacuum tubes that produces a narrow, collimated electron beam that has a precise kinetic energy. & electron\\
        \hline
        Vala (programming language) & Comp. & Vala is an ... in May 2006.&Vala is an object-oriented programming language with a self-hosting compiler that generates C code and uses the GObject system. &compiler\\
    \bottomrule
    \end{tabular}
    \caption{Snapshot of the \datawiki dataset}
    \label{tab:appendix:wikidomains_snapshot}
\end{table*}

\subsection{Difficult concept statistics}
In roughly 85\% of \datawiki examples, we extracted the \difficultconcept using a ratio of how often the concept appears within this domain compared to Wikipedia overall (Equation~\ref{eq:domainspec}).  For those examples, the average computed ratio for the selected \difficultconcept is 0.9. In the remaining 15\% of examples, the \difficultconcept was chosen using the age of acquisition lexicon \citep{kuperman2012age}.  On average, each \difficultconcept contains 1.3 tokens.

\section{Prompts}\label{sec:appendix:prompts}

We experimented with 4 candidate prompts for both $explain$ and $simplify$ prompts categories. Table~\ref{tab:appendix:candidate_prompts} outlines these candidate prompts. To identify the best prompt, we applied the prompts to a set of 100 randomly sampled definitions from the \datawiki and \scidef datasets, and the lead author manually assessed the goodness of the rewrites, assigning a binary label 1 or 0 to each of the rewrites, indicating whether the rewrite successfully addresses the concept difficult or not respectively. The prompts that we use in our experimental setup had the highest number of definitions where the rewrite was assessed as a good rewrite.

\begin{table*}[h]
    \centering
    \begin{tabular}{lp{13cm}}
        \toprule
        Prompt Strategy & Prompt text \\
        \midrule
        simplify & \begin{tabular}[c]{@{}l@{}}Rewrite the definition simplifying the concept: ``cerebellum.''\end{tabular}\\
        simplify & Rewrite the definition making the concept simpler: ``cerebellum.''\\
        simplify & Rewrite the definition making the concept simpler: ``cerebellum.''\\
        simplify & Rewrite the definition simplifying difficulty with the concept: ``cerebellum.''\\
        \midrule
        explain & \begin{tabular}[c]{@{}l@{}}Rewrite the definition integrating an explanation for the concept: ``cerebellum.''\end{tabular}\\
        explain & Rewrite the definition adding an explanation for the concept: ``cerebellum.''\\
        explain & Rewrite the definition providing an explanation of the concept: ``cerebellum.''\\
        explain & Rewrite the definition to add content that explains the concept: ``cerebellum.''\\
        \bottomrule
    \end{tabular}
    \caption{Candidate prompts that we explored}
    \label{tab:appendix:candidate_prompts}
\end{table*}

Table~\ref{tab:prompts} details the ``simplify'' and ``explain'' prompts that we used in our study.

\begin{table*}[h]
    \centering
    \begin{tabular}{lp{13cm}}
        \toprule
        Prompt Strategy & Prompt text \\
        \midrule
        simplify & \begin{tabular}[c]{@{}l@{}}Rewrite the definition simplifying the concept: ``cerebellum''.\\
        Definition: Chiari malformations (CMs) are structural defects in the cerebellum.\\
        Rewrite:\end{tabular}\\
        \midrule
        explain & \begin{tabular}[c]{@{}l@{}}Rewrite the definition integrating an explanation for the concept: ``cerebellum''.\\Definition: Chiari malformations (CMs) are structural defects in the cerebellum.\\Rewrite:\end{tabular}\\

        \bottomrule
    \end{tabular}
    \caption{Prompts used in the experimental evaluation}
    \label{tab:prompts}
\end{table*}

\section{Instructions for human evaluation}
\label{sec:appendix:userstudy_llm_evals}
To evaluate LLM-generated definitions for their suitability for simplifying \domain specific concepts, we show a human rater the following 1) the original definition, 2) a difficult concept $c^d$ within the definition that we identified, 3) the LLM-rewritten definitions. We ask raters to answer the following 1) Please rate on a scale of 1-5 how much the REWRITE preserves the meaning of the original, 2) Can someone understand the definition if they do not know the difficult concept: X? (Yes/No), 3) Please rate which of the ORIGINAL and REWRITE are easier to understand? (Original/Rewrite/Both), 4) Please rate your level of familiarity with the concept.

\textbf{Rationale for human evaluation questions}
\label{sec:appendix:human_eval_rationale}
We cannot control readers' familiarity with the concept, therefore we rely on their understanding to determine someone's ability to understand the definition without knowledge of the \difficultconcept. How much is a definition understandable to someone is dependent on their background knowledge. Therefore, by asking whether the REWRITE is easier to understand or the ORIGINAL definition, we rely on the annotator's opinion of whether the rewritten definition gives them a better understanding of the topic.

\begin{figure}
    \centering
    \includegraphics[width=\linewidth]{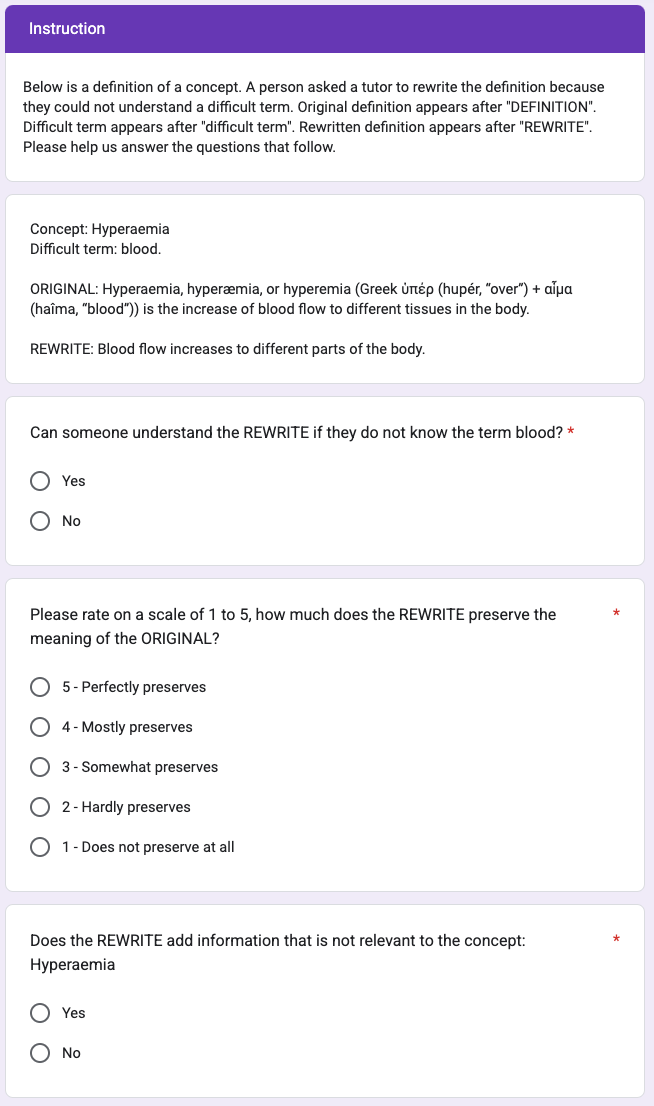}
    \caption{Annotation example for evaluating LLM rewritten definitions for concept simplification.}
    \label{fig:study_evaluation_example}
\end{figure}

Each annotator was presented with about 15 definitions to answer questions about, and the total annotation time per annotator was about 20-25 minutes. Before the annotation, we briefed the annotators about task and provided two examples to help them understand the task of concept simplification. Figure~\ref{fig:study_evaluation_example} shows screenshot of the annotation task.

We displayed a consent form to the participants detailing the study and that the risks would be no more than assessing definitions written by AI and gave them the option to leave the study at any time. We compensated the participants above the hourly minimum wage based on their demographic location. The study was approved by the internal ethics review team.

We collected the educational background of annotators, summarized in Table~\ref{tab:appendix:demographic_bg}.

\begin{table}[h]
    \centering
    \begin{tabular}{lc}
    \toprule
        Background & Percentage \\
    \midrule
        4-year college degree &               48\%\\
        Master's degree &                      16\%\\
        2-year college degree &                14\%\\
        Some college &                         13\%\\
        High school &                           3\%\\
        Professional degree (MD, JD, etc) &     2\%\\
        Doctoral degree (PhD) &                 0.8\%\\
    \bottomrule
    \end{tabular}
    \caption{Educational background of annotators for human evaluations of LLM-rewrites. Total number of annotators was 229.}
    \label{tab:appendix:demographic_bg}
\end{table}

\section{Inference setting}
For open-source models, we run inference on GPUs using the Huggingface\footnote{\url{huggingface.co}} transformers implementation. To fit Falcon and Bloom models on the available GPUs, we run the models with 8-bit quantization~\cite{dettmers2022llm}. 
For the commercial models, we use the publicly available APIs to query the models and generate outputs. For all LLMs, we use top-$k$ sampling\footnote{We set the value of $k$ to 40}.

\section{Results of Zero-shot Prompting}
\label{appendix:sec:zerofew}
See Table~\ref{tab:zerofew} for the zero-shot performance results.  As expected, scores are generally lower with zero-shot than few-shot. In particular, ICL examples seem to help with the meaning preservation dimension pretty consistently.

\begin{table}[tb]
    \centering
    \resizebox{0.75\linewidth}{!}{%
    \begin{tabular}{cllrrr}

    \toprule
         & & {LLM}  & {\hmp} & {\hre} & {\hru} \\
    \midrule

\multirow{8}{*}{\rotatebox[origin=c]{90}{\small \sc WikiDomains}} &	\multirow{4}{*}{\rotatebox[origin=c]{90}{\small simplify}}	&	Bloomz	&4.12	&	0.23	&	0.48	\\
                                                            &	&	Falcon	&3.88	&	0.53	&	0.78	\\
                                                            &	&	PaLM2	&4.27	&	0.08	&	0.60	\\
                                                            &	&	GPT4	&4.02	&	0.67	&	0.70	\\
\cmidrule{2-6}
&	\multirow{4}{*}{\rotatebox[origin=c]{90}{\small explain}}	&	Bloomz	&4.23	&	0.18	&	0.41	\\
&		                                                        &	Falcon	&4.00	&	0.47	&	0.65	\\
&		                                                        &	PaLM2	&4.14	&	0.26	&	0.63	\\
&		                                                        &	GPT4	&4.11	&	0.72	&	0.89	\\

        \midrule
        \midrule
        \multirow{8}{*}{\rotatebox[origin=c]{90}{\small \scidef}} &	\multirow{4}{*}{\rotatebox[origin=c]{90}{\small simplify}}
        &	Bloomz	&4.01	&	0.17	&	0.56	\\
&		&	Falcon	&3.87	&	0.49	&	0.56	\\
&		&	PaLM2	&4.21	&	0.05	&	0.68	\\
&		&	GPT4	&4.36	&	0.47	&	0.93	\\
\cmidrule{2-6}
&	\multirow{4}{*}{\rotatebox[origin=c]{90}{\small explain}}
        &	Bloomz	&4.73	&	0.03	&	0.38	\\
&		&	Falcon	&3.72	&	0.47	&	0.63	\\
&		&	PaLM2	&4.53	&	0.37	&	0.64	\\
&		&	GPT4	&4.34	&	0.53	&	0.83	\\
    \bottomrule
    \end{tabular}%
    }
    \caption{Human evaluations of LLM-generated rewrites for targeted concept simplification with zero-shot setting}
    \label{tab:zerofew}
\end{table}

\begin{table}[h]
    \centering
    \begin{tabular}{lcc}
    \toprule
    & \multicolumn{2}{c}{Krippendorff's Alpha}\\
       Metric & IAA & Ann vs. Majority \\
    \midrule
       \hmp & 0.31 & 0.70\\
       \hru & 0.21 & 0.65\\
       \hre & 0.25 & 0.62\\

    \bottomrule
    \end{tabular}
    \caption{Krippendorff's alpha scores for the human evaluations of meaning preservation (\hmp, an interval score out of 5), rewrite understanding (\hru, binary score), and rewrite easier (\hre, binary score).  We report coefficients between pairs of annotators (IAA = inter-annotator agreement) and also the agreement between individual annotation and the majority vote label for that example (Ann vs Majority).}
    \label{tab:appendix:ka}
\end{table}

\section{Human evaluation agreement}
Table~\ref{tab:appendix:ka} shows the human evaluation agreement scores for our study.
The inter-annotator alpha scores show  weak agreement (in the range between 0.2-0.3), which is somewhat expected due to the subjective nature of some evaluations.  In aggregating scores we use the majority vote between the three annotators (or the mean in the case of \hmp).  The Krippendorff's alpha between individual ratings and the majority vote falls in the range of 0.6-0.7, showing that individual ratings are generally closely aligned with the majority rating.

\section{Automatic Metric Performances}
In Table~\ref{tab:automated_metrics}, we present the model performances on different automatic metrics.
\begin{table*}[h]
    \centering


    \begin{tabular}{llrrrrrr}
    \toprule
         & {LLM} & {Density} & \bleufour & {BertSc} &
         {$\Delta$Len} & {$\Delta$AoA} & {$\Delta$Flesch}\\
    \midrule
        \multirow{4}{*}{\rotatebox[origin=c]{90}{\small \sc WikiDomains}}
            & Bloomz &11.78&	0.53&	0.89&	1.61&	-0.24&	7.30 \\
            & Falcon &6.62&	0.27&	0.80&	12.72&	4.44&	{\bf 9.10 }\\
            & PaLM2 &{\bf 19.57} &	{\bf 0.74}&	{\bf 0.92}&	4.06&	2.28&	1.72 \\
            & GPT4 &4.62&	0.24&	0.83&	{\bf 20.28} &	{\bf 5.77}&	8.50 \\
        \midrule
        \multirow{5}{*}{\rotatebox[origin=c]{90}{\small \scidef}}
        & Bloomz & 12.81&	0.62&	0.92&	2.35&	-0.02&	5.16\\
        & Falcon &3.67&	0.20&	0.76&	9.92&	{\bf 4.55}&	4.51\\
        & PaLM2 & {\bf 17.67} &	{\bf 0.80}&	{\bf 0.95}&	2.71&	0.16&	2.64\\
        & GPT4 & 4.24&	0.24&	0.83&	{\bf 17.85} & 3.45&	{\bf 7.41}\\
    \bottomrule
    \end{tabular}%
    \caption{Automated evaluations of concept simplifications on the full test set of \datawiki and \scidef}
    \label{tab:automated_metrics}
\end{table*}

\section{Qualitative examples}
Table \ref{tab:concept_simplification_examples} discusses several example outputs from the {\it simplify} prompt and identifies potential problems in some of the generated definitions.
\label{appendix:sec:examples}
\begin{table*}[h]
    \centering
    \begin{tabular}{p{5cm}p{5cm}p{5cm}}
    \toprule
    Notes & Original definition & Rewritten definition\\
    \midrule
       PaLM2, Difficult concept: zenith & Tropic Of Capricorn is the southernmost latitude of latitude on the Earth at which the Sun appears zenith overhead at its culmination, which lies approximately 23.4 degrees south of the Equator. & Tropic Of Capricorn is the southernmost latitude of latitude on the Earth at which the Sun appears directly overhead at its culmination, which lies approximately 23.4 degrees south of the Equator. \\
       \midrule
       GPT4, Difficult concept: zenith  \textcolor{blue}{Note: Correctly simplifies difficult concept but changes definition significantly.} & Tropic Of Capricorn is the southernmost latitude of latitude on the Earth at which the Sun appears zenith overhead at its culmination, which lies approximately 23.4 degrees south of the Equator. & The Tropic of Capricorn is the furthest point south on Earth where the Sun can be seen directly overhead at its highest point. This is roughly 23.4 degrees south of the Equator.\\
       \midrule
       PaLM2, Difficult concept: vegetation & Jungle is an area covered with dense vegetation dominated by large trees, often tropical. & Jungle is an area covered with dense plants dominated by large trees, often tropical.\\
       \midrule
       GPT4, Difficult concept: vegetation  \textcolor{blue}{Note: Correctly simplifies difficult concept but changes definition significantly.} & Jungle is an area covered with dense vegetation dominated by large trees, often tropical. & A jungle is a place filled with a lot of plants and big trees, usually found in hot regions.\\
       \midrule
       Bloomz, Difficult concept: linguistics \textcolor{blue}{Note: Makes no change.} & Prolog is is a logic programming language associated with artificial intelligence and computational linguistics. & Prolog is a logic programming language associated with artificial intelligence and computational linguistics.\\
       \midrule
       GPT4, Difficult concept: linguistics \textcolor{blue}{Note: Correctly simplifies difficult concept but changes definition significantly.} & Prolog is is a logic programming language associated with artificial intelligence and computational linguistics. & Prolog is a computer language used in artificial intelligence and the study of how computers can understand and use human language.\\
       \midrule
       PaLM2, Difficult concept: ovoid \textcolor{blue}{Note: model drops information.} & A coccus (plural cocci) is any bacterium that has a spherical, ovoid, or generally round shape. & A coccus (plural cocci) is any bacterium that has a spherical, or generally round shape.\\
       \midrule
       GPT4, Difficult concept: ovoid &  A coccus (plural cocci) is any bacterium that has a spherical, ovoid, or generally round shape. & A coccus (or cocci for many) is a type of bacterium that is shaped like a sphere, an egg, or is generally round.\\
      \bottomrule
    \end{tabular}
    \caption{LLM rewrite examples for $\mathit{simplify}$ prompts}
    \label{tab:concept_simplification_examples}
\end{table*}

\section{Qualitative examples from the \textit{simplify} vs. \textit{explain} prompts}
Table \ref{tab:simplify_context_examples} shows side-by-side examples of the rewritten definitions when the same model is prompted with the \textit{simplify} prompt versus the \textit{explain} prompt.
\label{appendix:sec:examples}
\begin{table*}[h]
    \centering
    \begin{tabular}{p{2cm}p{3.5cm}p{3.5cm}p{5cm}}
    \toprule
    model & original definition & \textit{simplify} definition & \textit{explain} definition\\
    \midrule
Falcon (0-shot) & Jungle is an area covered with dense \textbf{vegetation} dominated by large trees, often tropical. & The jungle is a lush and verdant forest characterized by towering trees and dense undergrowth. & The jungle is a vast expanse of land that is covered in dense vegetation, primarily consisting of tall trees and other plants. It is characterized by its humid climate and high levels of rainfall, which support the growth of a wide variety of flora and fauna.
 \\
 \midrule
PaLM2 (3-shot) & Chiari malformations (CMs) are structural defects in the \textbf{cerebellum}. & Chiari malformations (CMs) are structural defects in the brain. & Chiari malformations (CMs) are structural defects in the cerebellum. The cerebellum is a region of the brain that plays an important role in motor control. \\
\midrule
GPT4 (3-shot) & Hamlet is a small human settlement, variably defined as one the size of a town, village, or \textbf{parish} or as a smaller subdivision of or satellite entity to a larger settlement. & Hamlet is a small human settlement, variably defined as one the size of a town, village, or church district or as a smaller subdivision of or satellite entity to a larger settlement. & Hamlet is a small human settlement, variably defined as one the size of a town, village, or parish or as a smaller subdivision of or satellite entity to a larger settlement. A parish is a type of administrative division used by some churches and by some civil authorities.\\
\midrule
Bloomz (3-shot) & Arginase deficiency is an inherited disorder that causes the amino acid arginine (a building block of proteins) and \textbf{ammonia} to accumulate gradually in the blood. & Arginase deficiency is a rare disease in which the body is unable to process certain proteins properly. & Arginase deficiency is an inherited disorder that causes the amino acid arginine (a building block of proteins) and ammonia to accumulate gradually in the blood. Ammonia is the toxic by-product of protein metabolism. \\
      \bottomrule
    \end{tabular}
    \caption{Definitions rewritten with \textit{simplify} vs.\ \textit{explain} prompts. The difficult concept is in bold.}
    \label{tab:simplify_context_examples}
\end{table*}

\end{document}